\documentclass{article}

\usepackage{microtype}
\usepackage{booktabs} 
\usepackage{xspace}
\usepackage{multirow}
\usepackage{graphicx}
\usepackage{subcaption}
\usepackage{booktabs}
\usepackage{wrapfig}
\usepackage{booktabs} % for professional tables

\usepackage{hyperref}

\usepackage[accepted]{icml2024}

\usepackage{amsmath}
\usepackage{amssymb}
\usepackage{mathtools}
\usepackage{amsthm}
\usepackage{booktabs} 

\usepackage{algorithmic}
\usepackage{algorithm}
\usepackage[capitalize,noabbrev]{cleveref}
\theoremstyle{plain}

\theoremstyle{definition}

\theoremstyle{remark}

\newcommand{\ie}{\textit{i.e.,}\xspace}
\newcommand{\eg}{\textit{e.g.,}\xspace}

\newcommand{\aka}{\textit{a.k.a.}\xspace}
\newcommand{\modelname}{nanoLM\xspace}
\newcommand{\muscaling}{$\mu$Scaling\xspace}

\usepackage[textsize=tiny]{todonotes}

\begin{document}

\twocolumn[
\icmltitle{nanoLM: an Affordable LLM Pre-training Benchmark via Accurate Loss Prediction across Scales}




\begin{icmlauthorlist}
\icmlauthor{Yiqun Yao}{baai}
\icmlauthor{Siqi Fan}{baai,uestc}
\icmlauthor{Xiusheng Huang}{baai,ucas}
\icmlauthor{Xuezhi Fang}{baai}
\icmlauthor{Xiang Li}{baai}
\icmlauthor{Ziyi Ni}{baai,ucas}
\icmlauthor{Xin Jiang}{baai}
\icmlauthor{Xuying Meng}{ict}
\icmlauthor{Peng Han}{uestc}
\icmlauthor{Shuo Shang}{uestc}
\icmlauthor{Kang Liu}{ucas}
\icmlauthor{Aixin Sun}{ntu}
\icmlauthor{Yequan Wang}{baai}

\end{icmlauthorlist}

\icmlaffiliation{baai}{Beijing Academy of Artificial Intelligence}
\icmlaffiliation{uestc}{University of Electronic Science and Technology of China}
\icmlaffiliation{ucas}{Institute of Automation, Chinese Academy of Science}
\icmlaffiliation{ict}{Institute of Computing Technology, Chinese Academy of Sciences}
\icmlaffiliation{ntu}{Nanyang Technological University}

\icmlcorrespondingauthor{Yequan Wang}{tshwangyequan@gmail.com}

\icmlkeywords{Machine Learning, ICML}

\vskip 0.3in
]



\printAffiliationsAndNotice{}  % leave blank if no need to mention equal contribution

\begin{abstract}
As language models scale up, it becomes increasingly expensive to verify research ideas because conclusions on small models do not trivially transfer to large ones. A possible solution is to establish a generic system that accurately predicts certain metrics for large models without training them. Existing scaling laws require hyperparameter search on the largest models, limiting their predicative capability. In this paper, we present an approach (namely \muscaling) to predict the pre-training loss, based on our observations that Maximal Update Parametrization ($\mu$P) enables accurate fitting of scaling laws close to common loss basins in hyperparameter space. With \muscaling, different model designs can be compared on large scales by training only their smaller counterparts. 
Further, we introduce \modelname: an affordable LLM pre-training benchmark
that facilitates this new research paradigm.
With around \textbf{$14\%$} of the one-time pre-training cost, we can accurately forecast the loss for models up to
\textbf{52B}. 

Our goal with \modelname is to empower researchers with limited resources to reach meaningful conclusions on large models. We also aspire for our benchmark to serve as a bridge between the academic community and the industry.
\end{abstract}

\section{Introduction}

\begin{table}
\centering
\caption{Current LLMs. We summarize five popular Transformer models. GPT3 and MT-NLG have a model size exceeding 100B parameters, with training data around 300B tokens. In contrast, other models are smaller in size but trained on data surpassing a trillion tokens.}
\label{tab: current llm}
\begin{tabular}{lrc}
\hline Model & Params & Tokens \\
\hline 
GPT-3  & 175 B & 300 B \\
MT-NLG & 530 B & 270 B \\
Chinchilla & 70 B & 1.4 T \\
Llama 2  & 7/13/70 B & 2 T \\
Falcon & 7/40/180 B & 1.5/1 T \\
\hline
\end{tabular}

\end{table}

Large Language Models (LLMs) pre-trained on Web-scale data have demonstrated impressive performance on various downstream tasks under a variety of evaluation protocols such as zero-shot, few-shot, and fine-tuning. 

Modern LLMs are based on the Transformer architecture \cite{DBLP:conf/nips/VaswaniSPUJGKP17}, and can be trained with unsupervised objectives including causal language modeling \cite{DBLP:conf/nips/BrownMRSKDNSSAA20}, masked language modeling, among others \cite{DBLP:conf/icml/WangRHSCBLR22}. Since researches on scaling laws \cite{DBLP:journals/corr/abs-2001-08361, DBLP:journals/corr/abs-2203-15556} reveal the potential of improving model performance by increasing the total computation, the community have been scaling up both the model sizes and training data \cite{DBLP:conf/nips/BrownMRSKDNSSAA20, DBLP:journals/corr/abs-2201-11990, DBLP:journals/corr/abs-2307-09288, falcon40b}, as briefly summarized in Table \ref{tab: current llm}.

However, this trend makes it increasingly difficult for average researchers to verify research ideas or search for hyperparameters (HPs): first, conclusions achieved with small models do not trivially transfer to larger ones; on the other hand, directly training large models for multiple times is a costly endeavor. For instance, as reported by Llama-2 \cite{DBLP:journals/corr/abs-2307-09288}, time to train the 7B, 13B, and 70B models on roughly 2 trillion tokens is 184k, 368k, and 1.7M GPU hours with A100-80GB, respectively. An entry point as such prevents the community from swiftly making improvements that are reliable on large scales, and results in enormous waste of computational resources and intellectual efforts. Thus, it is necessary to establish a pipeline to compare different LLM structures, algorithms, and hyperparameters with limited computational resources (\ie small models), while making sure the results are instructive for \textit{any} model scale. This is the goal of this paper.

As a possible solution to this issue, the technical report of GPT-4 \cite{gpt4} showed that some behaviors of large models can be predicted before the training starts (with unpublished methods). In this paper, we start by proposing a method that yields accurate loss prediction, namely \muscaling ~(a compound word of $\mu$P  \cite{DBLP:journals/corr/abs-2203-03466} and Scaling Laws \cite{DBLP:journals/corr/abs-2001-08361}), with experimental results supporting its correctness. Based on \muscaling, we establish a new paradigm for meaningful research on large models without actually training them. Finally, we propose an affordable benchmark for LLM pre-training studies, namely \modelname, which facilities this research paradigm for the community.

\paragraph{Contributions.} 
We substantiate our contributions as follows:
\begin{itemize}
\item We propose \muscaling, a loss prediction method based on $\mu$P and modified scaling laws. For hyperparameters (HPs) in the common \textit{loss basins}, the training loss can be accurately predicted by a power-law function w.r.t. model sizes, which includes embedding sizes, in contrast to existing methods. This method requires searching for the optimal HP \textit{only once} to predict loss in arbitrary model scale.
\item  We unlock a new LLM study paradigm that can directly compare the loss for different model designs on large scales without direct training. We facilitate this paradigm by proposing \modelname, an affordable benchmark. \modelname is compatible with mainstream Transformer architectures, including decoder-only structures (\eg GPT, Llama), encoder-only structures (\eg BERT), and encoder-decoder structures (\eg T5), and supports data parallelism strategies (Section \ref{method}). For benchmark evaluation, we publicly release a pre-training dataset with 100B/400B/1T/2T tokens, chosen from existing sources and categorized into various specialized domains (Section \ref{data benchmark}).
\item \textit{Effectiveness}: Empirically, we successfully utilized our method to forecast the loss for 12-layer GPT, Llama, BERT, and T5 models on the C4 and MC4 dataset. 
For more expansive models, we experiment on GPT models with 32 and 64 layers, culminating in sizes of 26B and 52B, respectively. Results indicate that the actual loss remains predictable (Section \ref{fitting result}).
\item \textit{Efficiency}: By generating a series of small proxy models with sizes ranging from 38M to 3.4B, predicting loss using \muscaling incurs only $13.1\%, 14.2\%$ of the one-time pre-training cost for 26B and 52B models, respectively. 
This demonstrates that \modelname can help researchers make affordable and meaningful comparisons between different model designs and serve as a new benchmark for LLM study (Section \ref{efficiency and performance}).
\end{itemize}
To foster reproducibility, we will open-source all our code and data of \modelname benchmark. Part of our code is attached in ``Supplementary Material''.

\begin{figure*}[h]
  \centering
  \begin{subfigure}{0.31\linewidth}
    \centering
    \includegraphics[width=\linewidth]{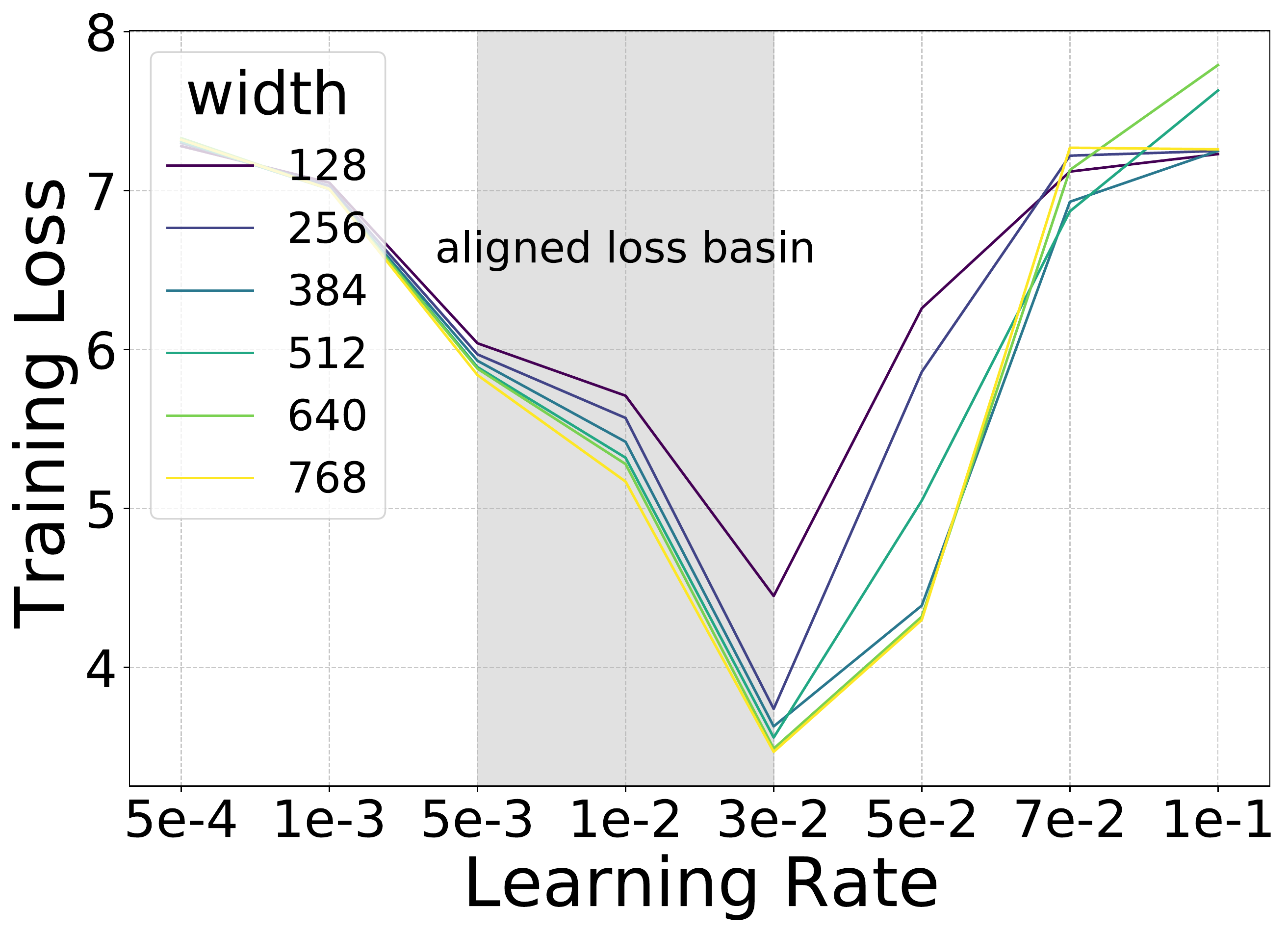}
    \caption{GPT.}
  \end{subfigure}
  \hspace{0.01\textwidth}
  \begin{subfigure}{0.31\linewidth}
    \centering
    \includegraphics[width=\linewidth]{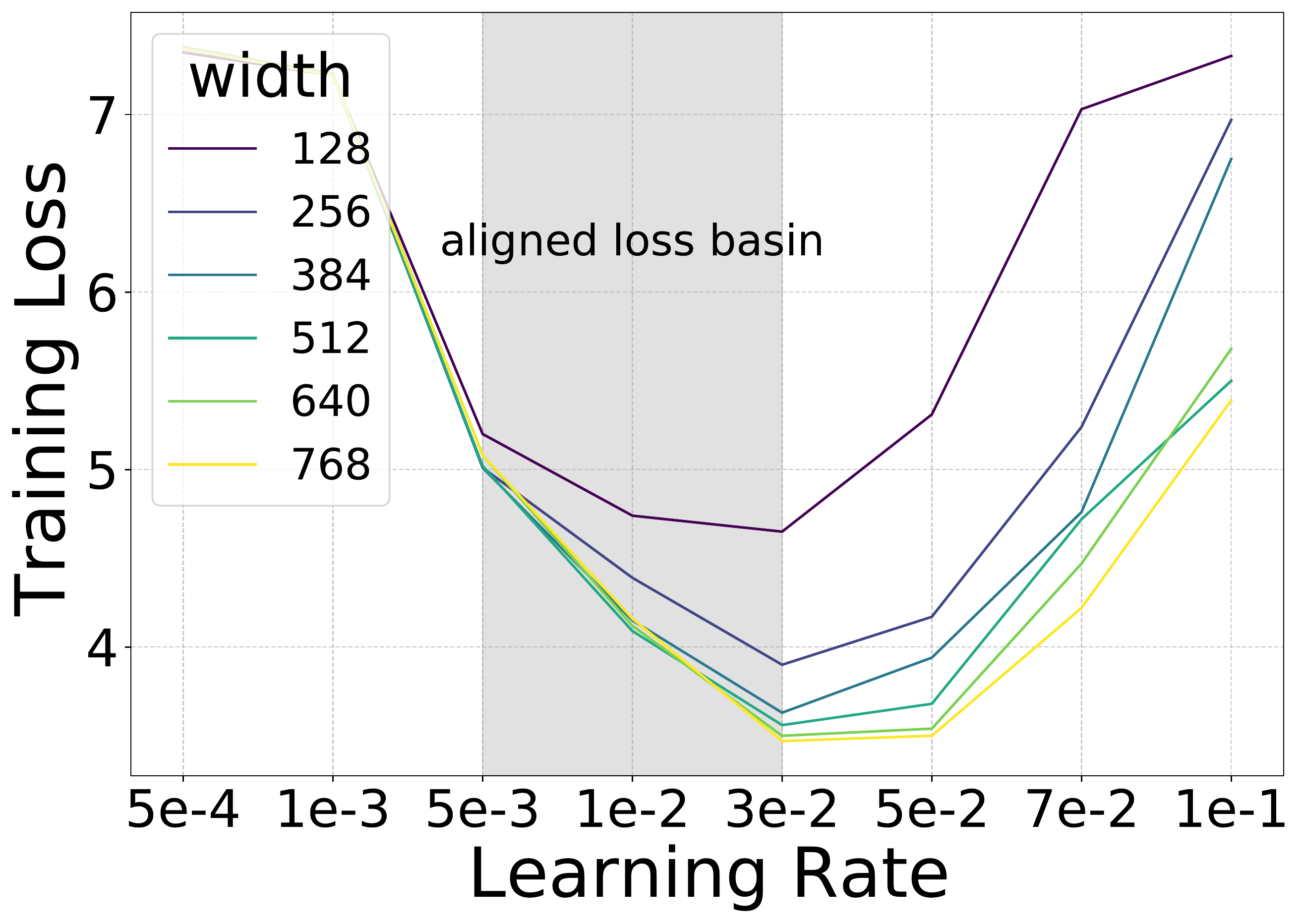}
    \caption{BERT.}
  \end{subfigure}
  \hspace{0.01\textwidth}
  \begin{subfigure}{0.31\linewidth}
  \centering
    \includegraphics[width=\linewidth]{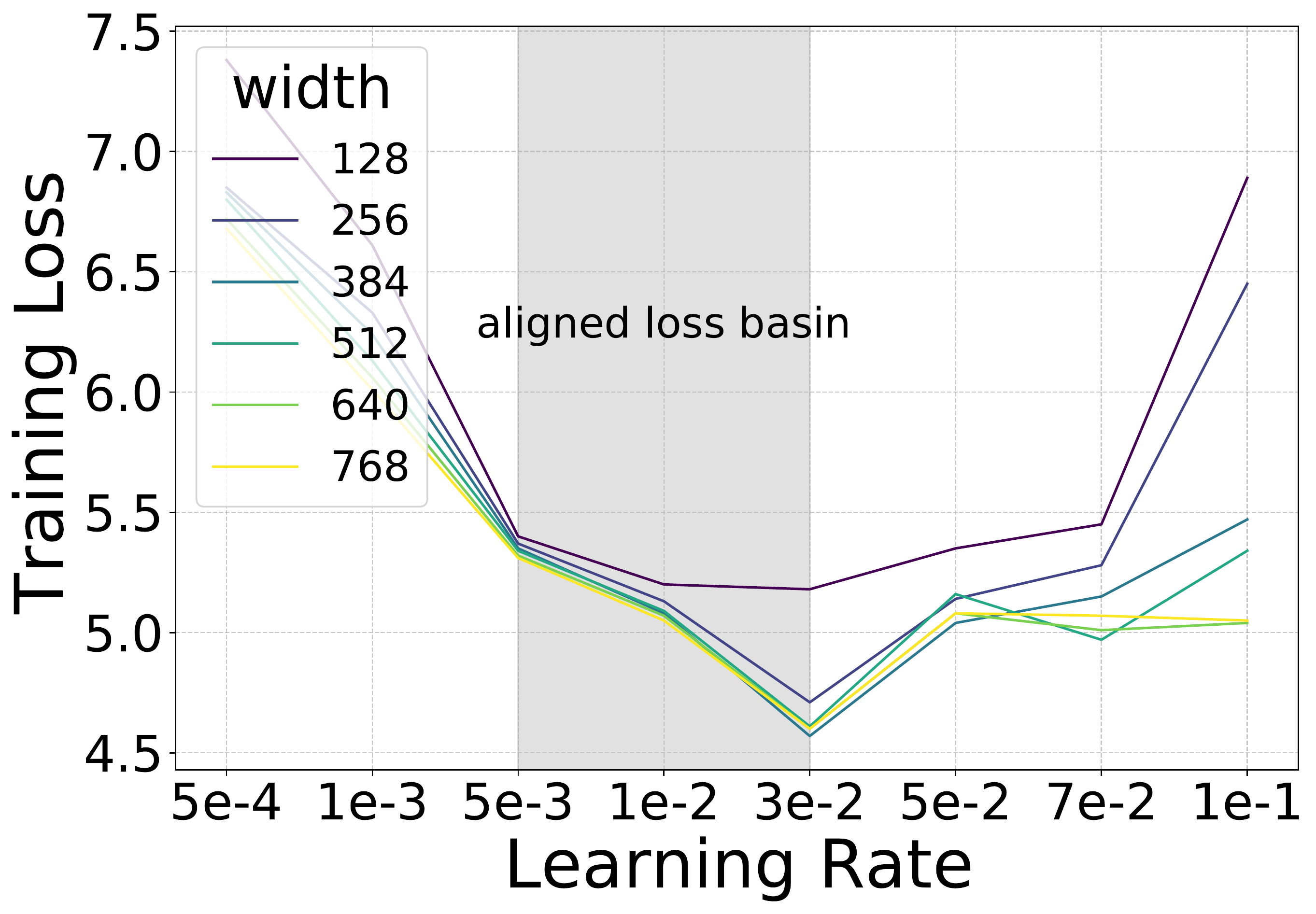}
    \caption{T5.}
  \end{subfigure}
  \caption{Pre-training loss w.r.t. learning rate for different Transformer architectures (\ie GPT, BERT, T5) across widths under $\mu$P. The result shows that the loss landscapes are aligned for models with different widths.}
  \label{fig: loss landscape}
\end{figure*}

\section{Preliminaries}
We first introduce our task setting (Section \ref{sec: setting}), and then discuss the pros and cons of current hyperparameter estimation methods (Section \ref{sec: mup}) and scaling laws (Section \ref{sec: scaling law}), as well as other related work (Section \ref{sec: related work}).
\subsection{Task Setting}
\label{sec: setting}
Our task is to do research with small models and safely generalize the conclusions to large-scaled models. This is not a trivial problem. Suppose a small model $M_{1}$ can be trained with an optimal HP $H_{1}$ to achieve a loss $L_{1}$, while research suggests another \textit{model design}, $M_{2}$, has a better $L_{2}\!<\!L_{1}$ with hyperparameter $H_{2}$. Then, for a LARGER version of both models ($M_{1}'$ and $M_{2}'$), typically, we have $H_{1}'\!\neq\! H_{1}$ and $H_{2}'\!\neq\! H_{2}$. Thus, both $H_{1}'$ and $H_{2}'$ must be re-searched by training the large models. Moreover, there is also no guarantee that $L_{2}'\!<\!L_{1}'$ still holds. We use these notations throughout our paper.

\subsection{Estimating HPs for Large Models}
\label{sec: mup}
Maximal Update Parametrization ($\mu$P, \cite{DBLP:journals/corr/abs-2203-03466}) is a transferring function for certain classes of HPs (i.e. $\mu$Transferable HPs, including learning rate, initialization variance, and multipliers) w.r.t. model widths (e.g. the hidden size of Transformers). 
Theories \cite{nongaussian, adaptive, tp4} suggest that for two models with the only difference being their widths $w$ and $w'$, their \textit{optimal} $\mu$Transferable HPs $H$ and $H'$ satisfy $H'\! =\! \mu P(H,r)$, where $r\!=\!w'/w$. In other words, under such transfer function, the loss landscape (w.r.t. HPs) of models in different widths are \textit{aligned}. See Figure \ref{fig: loss landscape} for a demonstration based on our reproduction. 
Please refer to the original papers for theoretical derivations.

For the $\mu$Transferable HPs, $\mu$P enables grid-searching on small models and directly transferring to large ones. However, it only provides insights on which HPs are better, but can \textit{not} predict the loss values themselves under these HPs. Meanwhile, may studies actually care about different model structures (e.g. GPT vs. Llama) and non-$\mu$Transferable HPs (e.g. layer numbers). Since $\mu$P does not cover these cases, there are very limited reliable approaches to know ``how many layers should I use for a 100B model'' unless we could somehow know the final training loss for each layer number. These cases are what we actually mean by \textit{model designs} in Section \ref{sec: setting} and elsewhere.

\subsection{Estimating Loss for Large Models}
\label{sec: scaling law}

The term “scaling laws” in modern deep learning stands for the relations between the performance of models (usually the test loss or some other performance metrics) and computational scales (like model size, dataset size, or training compute) \cite{DBLP:journals/corr/abs-1712-00409, DBLP:journals/corr/abs-2001-08361, epoch2023scalinglawsliteraturereview, DBLP:conf/iclr/RosenfeldRBS20}. 
These relations give researchers insight on trade-offs between model capabilities and computational budgets. However, these relations must be established through HP tuning across all model scales. Often, the optimal HP choices are unknown before the training of large models finish. Thus, existing scaling laws have limited power for ``loss prediction''.

\subsection{Other Related Work}
\label{sec: related work}
Some existing work explores HP transfer learning among different tasks or datasets \cite{DBLP:conf/nips/PerroneJSA18,DBLP:journals/corr/abs-2010-13117,DBLP:conf/icml/SalinasSP20}. In contrast, we focus on implementing loss prediction via HP transfer across different model widths, under the same task or data. 

Our \modelname benchmark is also related to current evaluation datasets for LLMs \cite{bigbench,mmlu}. A critical difference is that \modelname does not aim at directly evaluating pre-trained model checkpoints, but rather encourages the users to pre-train new models from scratch, and use the pre-training loss as metric for comparison. Thus, \modelname contains pre-training code with implementations of our proposed \muscaling on different models, in addition to data processing functions. 
Pre-training on nanoLM is made affordable via loss prediction, which avoids running the largest models. This is orthogonal to the optimization of parallelism mechanics and hardware usage \cite{DBLP:journals/corr/abs-1909-08053,dao2023flashattention, deepspeed}.

\begin{table}[ht]
    \centering
    \caption{$\mu$P function for a model $M^{\prime}$ that is r times the widths of M. If a parameter tensor has 2 dimensions that goes infinite when the model width goes infinite, it is “matrix-like” (\eg a fully-connected hidden layer); if the number is 1 or 0, it belongs to the “others” class. Note that embedding layers are “others”. “Output” means the layer that maps an infinite dimension to a finite dimension, which is the word decoding layer ($lm\_head$) in Transformers. A multiplier is a constant multiplied by a parameter tensor, which has a similar function to softmax temperature.}
    \begin{tabular}{c|cc}
    \hline Hyperparameter (weight) & $M$ & $M^{\prime} \sim r$ \\
    \hline AdamW learning rate (matrix-like) & $l$ & $l / r$ \\
    AdamW learning rate (others) & $l$ & $l$ \\
    Initialization variance (matrix-like) & $\sigma$ & $\sigma / r$ \\
    Initialization variance (others) & $\sigma$ & $\sigma$ \\
    Multiplier (output) & $\tau$ & $\tau / r$ \\
    Multiplier (others) & $\tau$ & $\tau$ \\
    \hline
    \end{tabular}
    \label{tab:muP}
\end{table}

\section{\modelname with \muscaling}

\begin{figure*}
  \centering
    \includegraphics[width=0.8\linewidth]{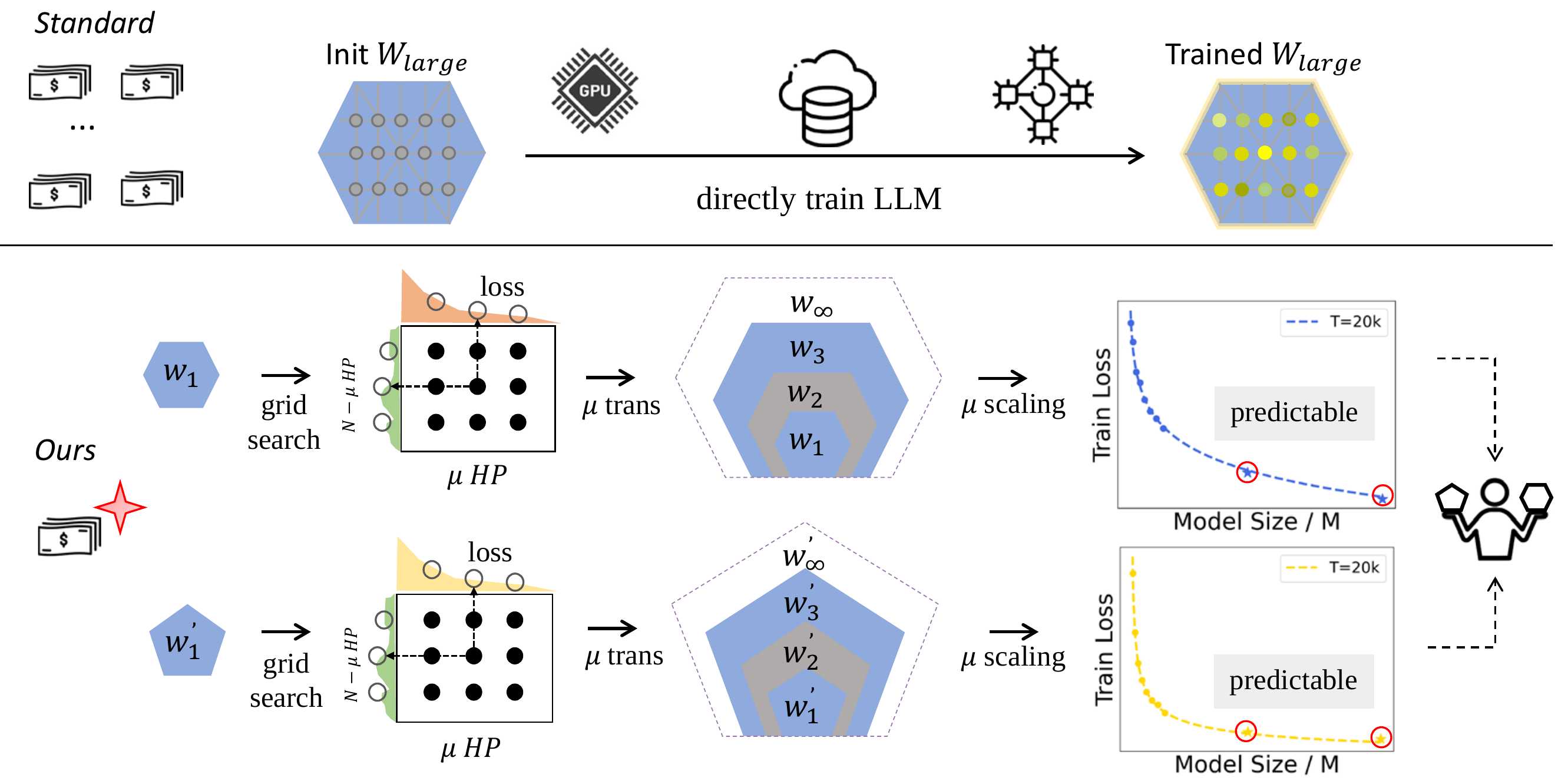}
    \caption{Illustration of standard LLM pre-training vs. \modelname. \textbf{Top}: Directly pre-training LLM with high computational cost, large data, and distributed training. \textbf{Down}: loss prediction with \muscaling. In this scenario, two different model designs, $M$ and $M'$, are being compared. The process unfolds into four phases: 1) Shrink the width of the two models to base dimensions, denoted by $w_1$ and $w_1^{\prime}$, for grid search on $\mu$Transferable HPs. 2) Choose a series of proxy models with small widths, then apply $\mu$P for zero-shot HP Transfer for each model. 3) Train these proxy models, record the losses, and fit the scaling law. 4) Directly predict the loss at any given width without training large LLMs.}
    \label{overview}
\end{figure*}

In this section, we introduce our generic benchmark for LLM study without full-scaled training, namely \modelname. Figure \ref{overview} provides the overview of our benchmark. Section \ref{method} contains details of our \muscaling method and the inducted new research paradigm, and Section \ref{perform} introduces the components of our benchmark.

\subsection{Unlocking a New LLM Research Paradigm} \label{method}
\paragraph{Pre-training Loss as an Indicator of Model Performance.}
LLMs' training involves cross-entropy objectives such as predicting subsequent or masked tokens. Since typically the training data is processed for less than 1 epoch, this loss also strongly correlates with the perplexity on validation set. Besides, recent researches \cite{DBLP:journals/corr/abs-2307-09288,schaeffer2023emergent} suggest a link between pre-training loss and downstream task performance, although this is challenging to assess comprehensively. In short, for the same data, pre-training loss is a reasonable indicator of model capabilities.

\paragraph{\muscaling for Loss Prediction.}
We explore the problem of predicting the loss of a target (large) model with a series of small \textit{proxy} models. Based on the discussions in Sections \ref{sec: mup} and \ref{sec: scaling law}, we observe that in this problem setting, $\mu$P and scaling laws are complementary: first, scaling law offers a formal function to directly calculate the loss values with model scales, but the optimal HP of each model should be searched separately; second, $\mu$P provides a unified HP for multiple models varying only in width, but does not guarantee the exact loss values except knowing ``the wider, the better''. Thus, we develop a loss prediction approach, namely \muscaling, as follows (Figure \ref{overview}, bottom): 

First, we generate a series of proxy models $\{M(w_{0}), M(w_{1}), M(w_{2}), ..., M(w_{n})\}$ with $M_B\!=\!M(w_{0})$ being the base model and $M_T\!=\!M(w_{n})$ being the target large model. They differ only in widths $\{w_{0}, w_{1}, ..., w_{n}\}$. 

Second, we grid-search for the optimal $\mu$Transferable HPs $H$ = (learning rate, init variance, multipliers) on $M_B$. $\mu$P concludes that $H(i)\!=\!\mu P(H(0), w_{i}/w_{0})$ are also the optimal HPs for each $M(w_{i}), i\!\in\! [0,n]$. We illustrate the $\mu$P function we use in Table \ref{tab:muP}, which corresponds to Table 8 in \cite{DBLP:journals/corr/abs-2203-03466}.

Third, we select some small $w_{i}$-s, train the models $M(w_{i})$ with $H(i)$, and achieve loss $L(i)$. Based on these points, we fit a power-law function $L(C)=aC^{b}+c$ w.r.t. the number of parameters $C$, and expect that this function accurately predicts $L_T\!=\!L(n)$ that could be achieved by training $M_T\!=\!M(w_{n})$ with HP $H_T\!=\!H(n)$. 

The correctness of the third step is critical and has not been studied yet, but we show by our comprehensive experiments (Section \ref{perform}) that it is effective for HPs within the common \textit{loss basins} in the HP space, and with our modification on scaling laws (below). Note that the formulation of scaling law is empirical and \textit{should be} validated by experiments \cite{DBLP:journals/corr/abs-2001-08361,DBLP:journals/corr/abs-2203-15556,gpt4}.

\textit{Embedding counts as model size.} An interesting difference between \muscaling and mainstream scaling laws is we find that \muscaling fits better if the size of \textit{embedding} is counted in the model size, while methods like \cite{DBLP:journals/corr/abs-2001-08361} benefit from excluding it. We explain this in Section \ref{anal_studies}.

\paragraph{New Paradigm: Research without Re-search.}
Based on \muscaling, the $\mu$Transferable HPs are actually ``tunnels'' towards directly comparing different \textit{model designs} on large scales via loss prediction. For each model, we predict its loss via \muscaling by shrinking its width into a series of small ones and do not need to actually train the large model with full width. In this process, only one HP search is needed for each model design. We formalize this new paradigm in Algorithm \ref{alg:algorithm}, and highlight its feature as ``research without re-search''. This makes LLM studies more affordable, while still reaching meaningful conclusions for model scales beyond 50B.

\begin{algorithm}[tb]
  $\mathcal{H}$: all combinations of $\mu$Transferable HPs; $\mathcal{W}$: all possible widths;
  $\mathcal{P}$: all possible \textit{model designs}, which are the objectives of \emph{research}.\\
  $p^{*}\in\mathcal{P}$: the best model found;
  $h^{*}\in\mathcal{H}$: the optimal $\mu$Transferable HPs;
  $w_{T}\in\mathcal{W}$: the target width to predict;
  $\mathcal{D}$: training data.
  \begin{algorithmic}[1]
  \FOR{$p$ in $\mathcal{P}$}{
  \STATE Select the base and target widths $w_{0}, w_{T} \in \mathcal{W}$;
  \STATE $h_{0}^{*} \leftarrow grid\_search(w_{0}, p, \mathcal{H})$;
  \FOR{some small $w_{i}$ in $\mathcal{W}$}
  \STATE $h_{i}^{*}\leftarrow \mu P(h_{0}^{*}, w_{i}/w_{0})$;
  \STATE $L_{i}\leftarrow train(w_{i}, p, h_{i}^{*}, \mathcal{D})$;
  \STATE $C_{i}\leftarrow count\_params(w_{i})$;
  \ENDFOR\\
  \STATE Fit $L=aC^{b}+c$ with all ($L_{i}$, $C_{i}$)s;
  \STATE $C_{T}\leftarrow count\_params(w_{T})$;
  \STATE $L_{T}(p)\leftarrow aC_{T}^{b}+c$; $\#$ Loss Prediction
  \STATE $h_{T}^{*}\leftarrow \mu P(h_{0}^{*}, w_{T}/w_{0})$;
  \IF{$L_{T}(p) < L(p^{*})$}
  \STATE $p^{*}\leftarrow p$;
  \ENDIF
  }
  \ENDFOR
  \STATE Conclusion: model $p^{*}$ is better according to the predicted loss on ($w_{T}$, $h_{T}^{*}$).
  \STATE Optional: $train(w_{T}, p^{*}, h_{T}^{*}, \mathcal{D})$.
  \end{algorithmic}
  \caption{New Paradigm: Research without Re-search}
  \label{alg:algorithm}
\end{algorithm}

\subsection{The \modelname benchmark} \label{data benchmark}
We introduce \modelname, an implementation of our research paradigm on different Transformer-based architectures, as well as a curated dataset serving as a benchmark for training and loss comparison.
\subsubsection{Supported Architectures.}
\modelname is based on PyTorch\footnote{\url{https://pytorch.org/}.} and supports three popular and remarkable architectures: decoder-only structures (\eg GPT \cite{DBLP:conf/nips/BrownMRSKDNSSAA20} and Llama \cite{DBLP:journals/corr/abs-2302-13971}), encoder-only structures (\eg BERT \cite{DBLP:conf/naacl/DevlinCLT19}), and encoder-decoder structures (\eg T5 \cite{DBLP:journals/jmlr/RaffelSRLNMZLL20}).
%each leaving an indelible mark in the field of natural language processing. 
Furthermore, considering the potential GPU memory overflow due to excessive model parameters and sequence lengths, \modelname integrates Fully Sharded Data Parallel (FSDP, \cite{DBLP:journals/corr/abs-2304-11277}), which shards the optimizer states, gradients, and parameters across multiple GPUs.

\subsubsection{Pre-training Data}
To facilitate researchers in using \modelname for comparative analysis across different model designs, 
we carefully construct a curated pre-training datasets from those of existing large-scale models (\ie Llama, Falcon, GPT-3). It covers diverse domains to improve the generalization capabilities of the resultant models.

\paragraph{Data Statistics.}
Our pre-training dataset, as detailed in Appendix Table \ref{tab:dataset}, comprises a rich blend of various open-source materials that span a wide range of domains. Largely, we repurpose data sources previously utilized for training other LLMs, adhering strictly to the criteria that the data must be publicly accessible and conducive to open-sourcing. Our training data contains 4 tracks with 100B, 400B, 1T, or 2T tokens, representing different settings of data scales. All versions are sampled according to the proportions in Table \ref{tab:dataset}.

\paragraph{Categorized Domains.}
We add widely-ranged domain data to increase the diversity. We categorize the domains as follows.

\emph{WebText:} We encompass Falcon RefinedWeb \cite{DBLP:journals/corr/abs-2306-01116}, a substantial English web dataset derived from CommonCrawl, featuring rigorous filtering and extensive deduplication. We also include OpenWebText2 \cite{pile}, a sizeable dataset of filtered text documents, collected from URLs in Reddit submissions.

\emph{Professional Knowledge:} We select Books3 \cite{pile}, which consists of a mix of fiction and nonfiction books.

\emph{World Knowledge:} We add English Wikipedia to our training dataset, which is a standard source of high-quality text for language modeling.

\emph{Code:} We include the public GitHub dataset and hope to improve downstream performance on code-related tasks. 

\emph{Academic:} For scientific knowledge, we include arXiv for the training dataset, which consists of preprint research papers \cite{DBLP:conf/nips/LewkowyczADDMRS22}. These papers are mainly in the fields of math, computer science, and physics.

\emph{Question Answering:} We include a dump of Stack Exchange, a website of high-quality questions and answers that covers diverse domains ranging from computer science to chemistry. 

\section{Experiments}\label{perform}
First, we outline our experimental setup in Section \ref{setting}. Then, we present empirical loss prediction results in Section \ref{fitting result}. Last, we delve into cost analysis in Section \ref{efficiency and performance}. For additional analysis, please refer to Section \ref{anal_studies} and Appendix \ref{appendix: more analy}.

\begin{table*}[ht]
    \centering
    \caption{Model parameters and loss across various widths and architectures. Note: loss values in \textbf{bold} represent predicted values.}
    \footnotesize
    \begin{subtable}{\linewidth}
    \centering
    \caption{12-Layer models: parameter count (in millions) and loss value.}
    \vspace{-3pt}
    \label{tab: num params for 12-layer}
    
    \begin{tabular}{c|cccccccccc}
        \toprule
        \multicolumn{2}{c}{Width} & 128 & 256 & 384 & 512 & 640 & 768 & 896 & 1024 \\
        \midrule
        \multirow{2}{*}{GPT} & Size / M & 8.82 & 22.36 & 40.61 & 63.59 & 91.28 & 123.69 & 160.82 & 202.67 \\
        & Loss@20k & 4.45 & 4.20 & 4.05 & 3.94 & 3.90 & 3.87 & 3.85 & 3.84 (\textbf{3.810}) \\
        \midrule
         \multirow{2}{*}{Llama} & Size / M & 9.59 & 25.47 & 47.64 & 76.10 & 110.85 & 151.90 & 199.24 & 252.86 \\
        & Loss@20k & 4.49 & 4.31 & 4.24 & 4.20 & 4.18 & 4.17 & 4.11 & 4.10 (\textbf{4.112}) \\
        \midrule
        \multirow{2}{*}{BERT} & Size / M & 22.52 & 51.28 & 86.33 & 127.67 & 175.31 & 229.24 & 289.45 & 355.96 \\
        & Loss@20k & 3.99 & 3.15 & 2.95 & 2.88 & 2.83 & 2.81 & 2.78 & 2.77 (\textbf{2.782}) \\
        \midrule
        \multirow{2}{*}{T5} & Size / M & 26.46 & 67.03 & 121.75 & 190.63 & 273.66 & 370.85 & 482.20 & 607.70 \\
        & Loss@20k & 4.75 & 4.66 & 4.60 & 4.58 & 4.52 & 4.47 & 4.42 & 4.40 (\textbf{4.415}) \\
        \bottomrule
    \end{tabular}
    \vspace{5pt}
    \end{subtable}
    \begin{subtable}{\linewidth}
    \centering
    \caption{32-layer and 64-layer models: parameter count (in billions) and loss value.}
    \vspace{-3pt}
    \label{tab: num params for 32-layer}
    \begin{tabular}{c|cccccccccc}
        \toprule
        \multicolumn{2}{c}{Width} & 256 & 384 & 512 & 640 & 768 & 896 & 1024 & 2048 & 8192 \\
        \midrule
        \multirow{2}{*}{32-layer GPT} & Size / B & 0.038 & 0.076 & 0.126 & 0.189 & 0.265 & 0.353 & 0.454 & 1.714 & 26.185 \\
        & Loss@7k & 3.92 & 3.76 & 3.65 & 3.59 & 3.54 & 3.49 & 3.47 & 3.45 & 3.41 (\textbf{3.381}) \\
        \midrule
        \multirow{2}{*}{64-layer GPT} & Size / B  & 0.077 & 0.153 & 0.254 & 0.381 &0.532&0.709&0.911&3.432& 52.385 \\
        & Loss@10k & 3.656 & 3.389 & 3.298 & 3.215 & 3.198 & 3.087 & 3.080 & 2.958 & 2.883 (\textbf{2.861}) \\
        \bottomrule
    \end{tabular}
    \vspace{5pt}
    \end{subtable}
    \label{tab:num params}
\end{table*}

\begin{figure*}[h]
  \centering
  \begin{subfigure}{0.45\linewidth}
    \centering
    \includegraphics[width=0.8\linewidth]{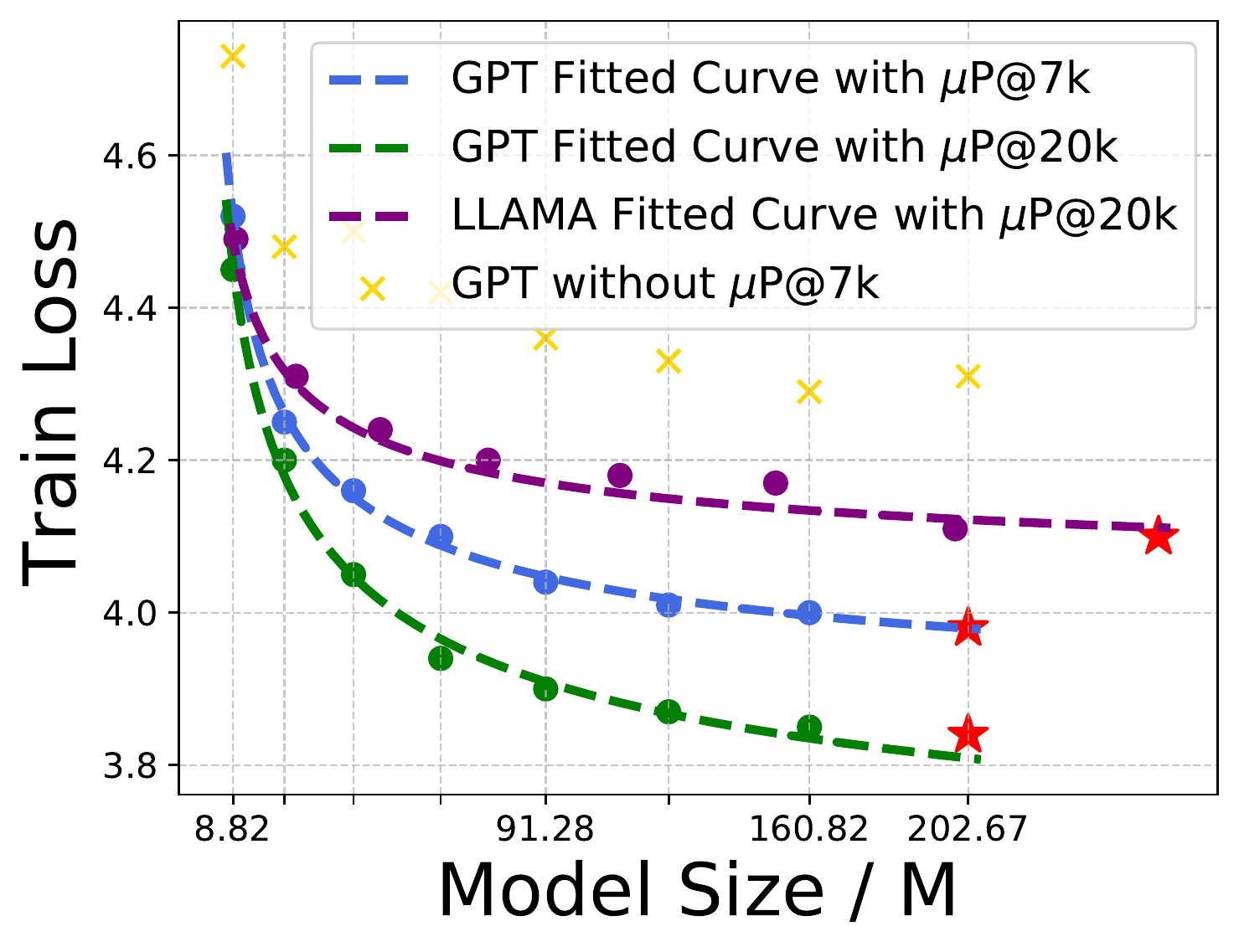}
    \caption{12-layer GPT and Llama.}
    \label{fig: 12-layer GPT}
  \end{subfigure}
  \hspace{0.01\textwidth}
  \begin{subfigure}{0.45\linewidth}
    \centering
    \includegraphics[width=0.8\linewidth]{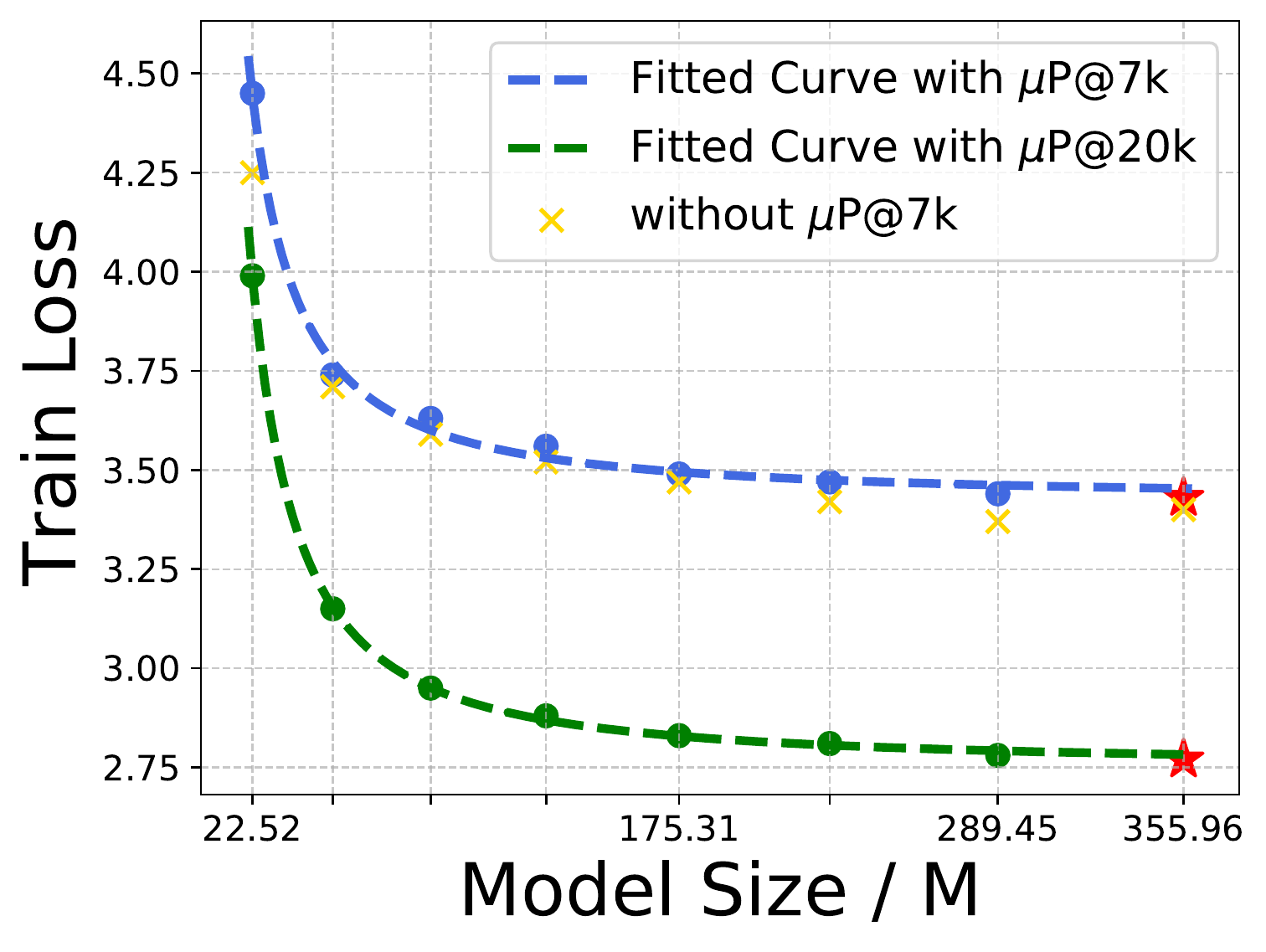}
    \caption{12-layer BERT.}
    \label{fig: 12-layer BERT}
  \end{subfigure}
  \hspace{0.01\textwidth}
  \begin{subfigure}{0.45\linewidth}
  \centering
    \includegraphics[width=0.8\linewidth]{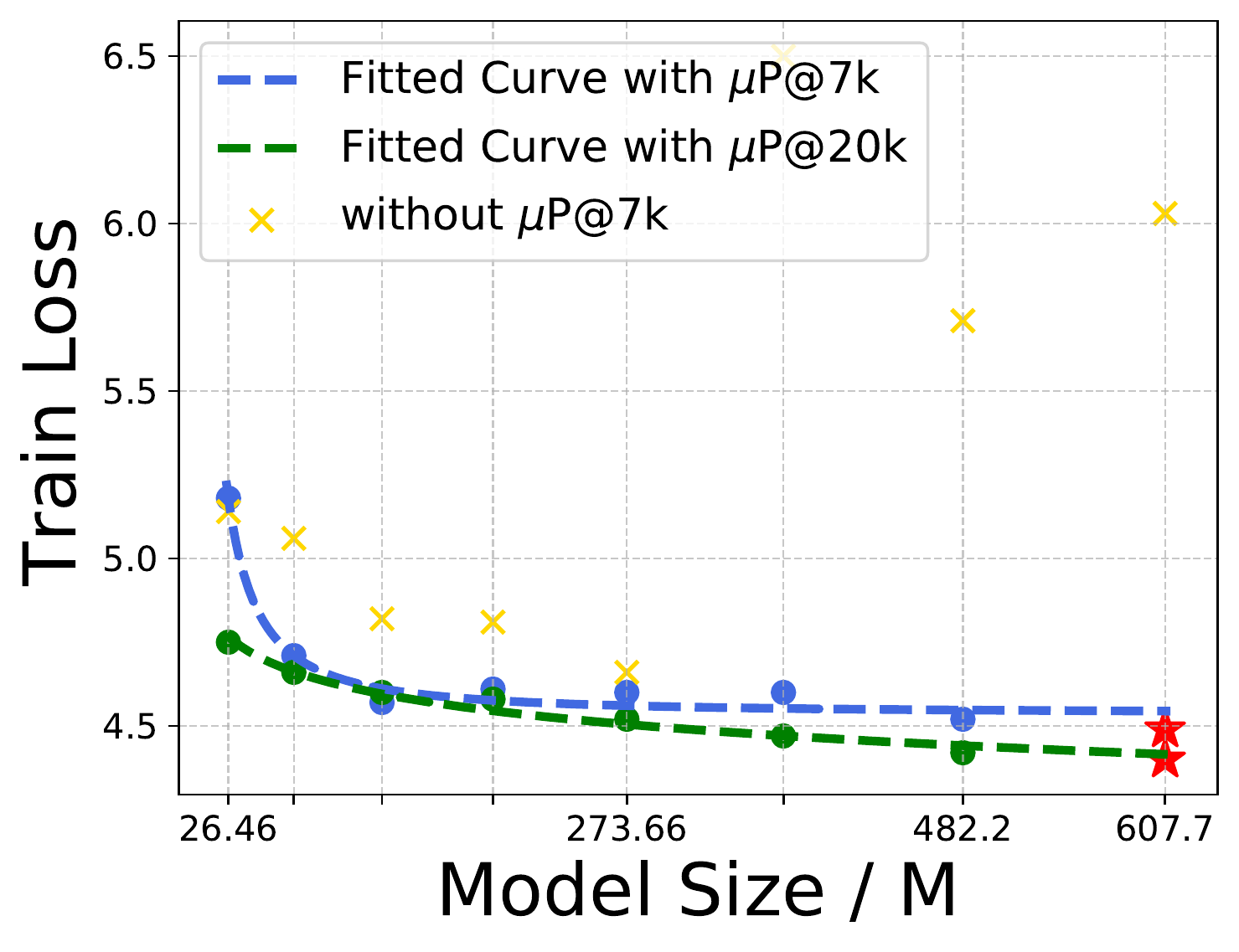}
    \caption{12-layer T5.}
    \label{fig: 12-layer T5}
  \end{subfigure}
  \begin{subfigure}{0.45\linewidth}
  \centering
    \includegraphics[width=0.8\linewidth]{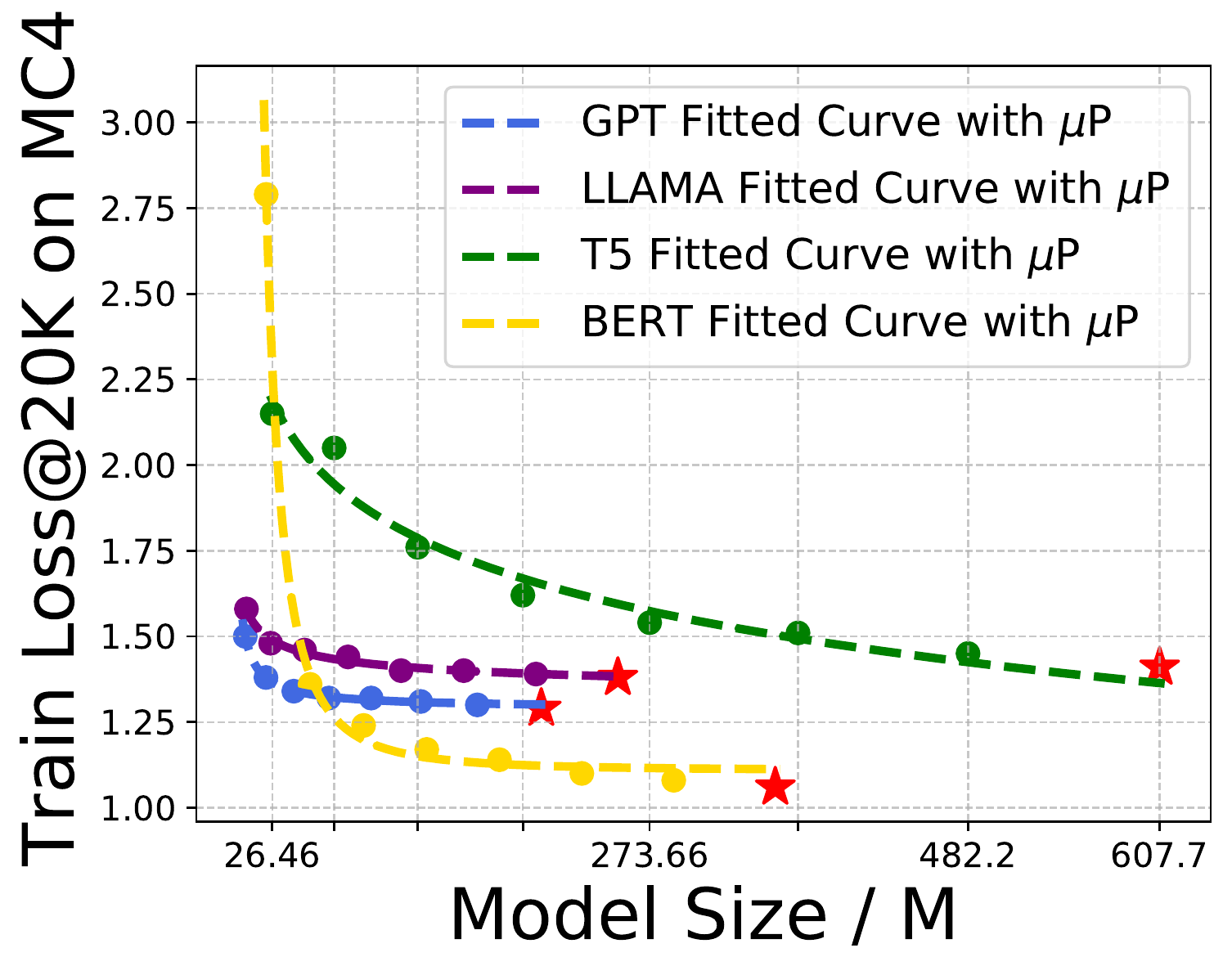}
    \caption{Training loss on MC4 dataset.}
    \label{fig: 12-layer mc4}
  \end{subfigure}

  \caption{Fitting result with $\mu$P and without $\mu$P: The dots illustrate the training loss across different small widths while incorporating $\mu$P. In contrast, the yellow cross points display the training loss at those very widths but without employing $\mu$P. We fit these dots to adapt the scaling law, aiming to ascertain if the loss of the final one models is consistent with this trend. The red star denotes the actual loss values from our training of the predicted wider models.}
  \label{fig: fitting results}
\end{figure*}

\subsection{Setup}\label{setting}
\paragraph{Model and Training Details.}
Our main goal is to justify the third step in \muscaling (Section \ref{method}), as well as the usability of the \modelname benchmark. To validate the loss prediction capability with different data scales, models, and computational resources, we experiment with the following three separate settings:

\textit{Single GPU.} Firstly, we test loss prediction on Transformer architectures (\ie GPT, BERT, T5) with 12 layers using the C4 \cite{DBLP:journals/jmlr/RaffelSRLNMZLL20} dataset under a single-GPU setup. We utilize a base width of 256 to conduct a grid search for the $\mu$Transferable HPs, primarily targeting the learning rate, with values ranging from $5e-4$ to $1e-1$\footnote{The $\mu$Transferable HPs include learning rate, initialization standard deviation, and multipliers. Due to computational resources, we focus on the learning rate. However, researchers can still explore the other $\mu$Transferable HPs on \modelname to achieve even better results.}. The number of parameters for the model series ranges from 8M to 700M with different widths, as detailed in Table \ref{tab: num params for 12-layer}. The target width for prediction is 1024 \footnote{Given that this group of experiments target on only one single GPU, both BERT and T5 ran out of memory when expanded to a width of 2048. In these cases, we leave the results for future.}. The batch size was established at 16, and the sequence length was set to 2048. According to \cite{DBLP:journals/corr/abs-2203-03466}, when the training steps surpass 5k, the $\mu$P transfer tend to stabilize. To this end, we choose to fit the loss at 7k and 20k steps and study its influences on the predicative performances. 

\textit{Multi-GPU with FSDP.} Secondly, to experiment with larger models and more data, We employ Fully Sharded Data Parallel (FSDP) \cite{DBLP:journals/corr/abs-2304-11277} to effectively address the training challenges. 
We conduct loss prediction with 32-layer GPTs on \modelname pre-training data (Section \ref{data benchmark}). We utilize a base width of 256 to conduct a grid search for the learning rate, ranging from $5e-4$ to $1e-1$. The number of parameters ranges from 38M to \textbf{26B} (Table \ref{tab: num params for 32-layer}). The batch size was established at 512, and the sequence length was set to 512. We use proxy models with widths under 2048 to predict the loss of the target model with width 8192.
    
\textit{Multiple Machines with Megatron.} Lastly, we further validate the feasibility of $\mu$P and $\mu$Scaling with extremely large widths. We conduct the experiments for the 64-layer GPTs on Megatron \cite{DBLP:conf/sc/NarayananSCLPKV21}. It provides efficient tensor, pipeline, and sequence-based model parallelism for pre-training LLMs. We utilize a base width of 256 to conduct a grid search for the learning rate, ranging from $1e-4$ to $1e-2$. The number of parameters ranges from 77M to \textbf{52B} (Table \ref{tab: num params for 32-layer}). The batch size was established at 512, and the sequence length was set to 2048. We exit training and fit the loss at 10k steps, with 10.49B tokens consumed (10$\%$ of the 100B track of \modelname pre-training data, Section \ref{data benchmark}).

All of our experiments are run on A100 GPUs. For each series of models used for loss prediction, the batches are fed into the models in the same order. More hyperparameters can be found in Appendix \ref{appendix: HP settings}.

\paragraph{$\mu$P and $\mu$Scaling Settings.}
All of our models are trained from scratch. Additionally, we adhere to the recommendation by \cite{DBLP:journals/corr/abs-2203-03466} to initialize the output word embeddings and query parameters as all-zero to avoid the Gaussian Process \footnote{Such Gaussian Process may cause misalignment of landscapes between small and large models.}. We maintain a consistent head dimension of 64 for each attention head and scale up the number of heads with model widths. The dropout and weight decay are set to 0. We employ the AdamW optimizer \cite{DBLP:conf/iclr/LoshchilovH19} with its default configurations. The ${a, b, c}$ coefficients in power law $L=aC^b+c$, as well as their standard deviations are computed with $scipy.optimize.curve\_fit()$ \footnote{\url{https://docs.scipy.org/doc/scipy/reference/generated/scipy.optimize.curve_fit.html}}.

\subsection{Main results: Fits $\&$ Extrapolation of \muscaling} \label{fitting result}

We use a base width of 256 to grid-search for the optimal $\mu$Transferable HPs. For 12-layer models, we found the optimal (learning rate, initialization standard deviation, output multipliers) being $(3e-2, 0.05, 5.0)$. For 32-layer and 64-layer GPTs, the optimal HPs are $(3e-2,0.04,4.0)$ and $(1e-3,0.01,1.0)$, respectively.
These HPs are $\mu$Transferred to other widths by Table \ref{tab:muP}. Some specific loss values can be found in Table \ref{tab:num params}, while the corresponding fitted curves are illustrated in Figure \ref{fig: fitting results}. For more numerical loss values, please refer to Appendix \ref{appendix: loss value}. The grid-search results can be found in Appendix \ref{appendix: GD search}.
\paragraph{12-layer Models on C4/MC4 with a Single GPU.} 
%\paragraph{Single-GPU Results.} 
For GPT, Llama, BERT, and T5, proxy models with widths $\leq$ 896 \footnote{We found a positive correlation between the number of data points and the accuracy of the fitted loss curve. We choose seven points due to the limited computation resource.} are used to fit the power-law curves and predict the loss for widths $>$ 896. For the 12-layer GPT and Llama, the fitted power-laws are $L=2.07*C^{-0.52}+3.85$ and $L=1.34*C^{-0.49}+4.02$, respectively. For 12-layer BERT, the result is 
$L=61.18*C^{-1.31}+3.43$; for the 12-layer T5, it is $L=0.25*C^{-0.47}+2.82$.
According to Figure \ref{fig: fitting results}, it's evident that based on the loss of the narrow proxy model series, we successfully predicted the loss of wider models. This holds for both 7k and 20k training steps and for all the 4 different Transformer variants, supporting the effectiveness of \muscaling. This extends the $\mu$P conclusion \cite{DBLP:journals/corr/abs-2203-03466} of ``the wider, the better" to ``wider is better adherent to a power law''. In contrast, for the yellow points, whose HPs are not transferred by $\mu$P but follows standard parametrization, the scaling law is not well fitted. They also underperform the same-width networks using $\mu$P.
Additionally, Figure \ref{fig: fitting results} (d) demonstrates that the conclusion still holds on diverse multilingual datasets, such as MC4 \cite{DBLP:journals/jmlr/RaffelSRLNMZLL20}.

\begin{figure}[t]
    \centering
     \includegraphics[width=0.8\linewidth]{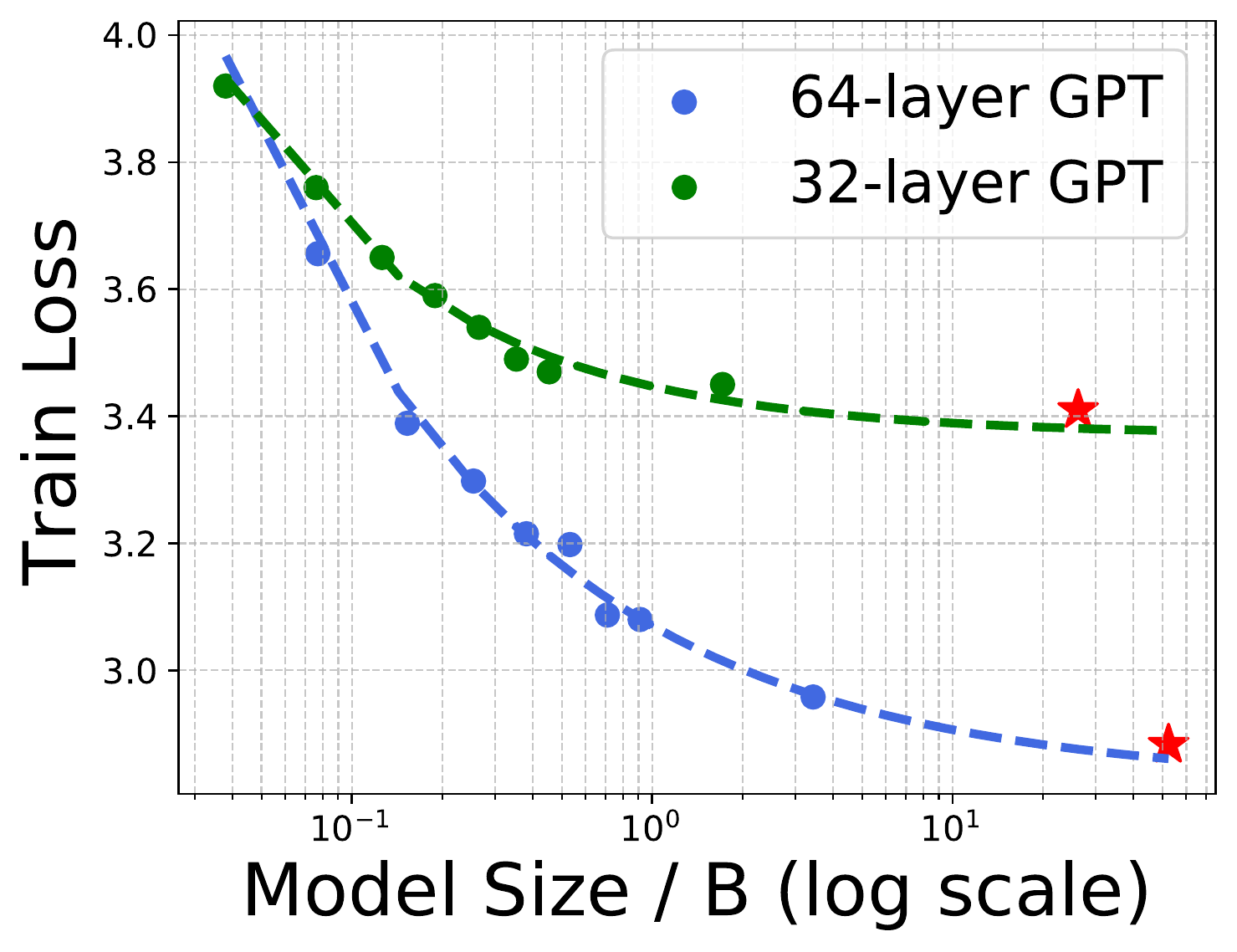}
    \caption{\muscaling results for large models. The blue/green dots signify the loss values of the proxy models. The red star denotes the actual loss values from our training of the target models.}
    \label{fig:32 and 64-layer GPT}
\end{figure}

\paragraph{32-layer GPTs with FSDP.} 
To validate \muscaling for larger-scaled models, we expanded the GPT to 32 layers. We train with the data parallel strategy (FSDP) on the benchmark pre-training data, as supported by \modelname (Section \ref{data benchmark}). Based on the proxy models with widths ranging from 256 to 2048, we predict the loss for a model with a width of 8192. The results are presented in Figure \ref{fig:32 and 64-layer GPT} with the power-law being $L=0.077*C^{-0.61}+3.37$. Notably, \muscaling successfully predicts the loss of a 26B model at 7k steps based on the losses from models ranging from 38M to 1.7B. This demonstrates that \modelname is an effective paradigm to study LLMs of billion-scale magnitude.

\paragraph{64-layer GPTs with Megatron.}
We further validate \modelname with extremely large models, exemplified by 64-layer GPTs implemented with Megatron \cite{DBLP:journals/corr/abs-1909-08053}. We train proxy models with widths ranging from 256 to 2048 using \modelname pre-training data and forecast the loss for a model in 8192 width. Results are presented in Figure \ref{fig:32 and 64-layer GPT} with the power-law being $L=0.249*C^{-0.47}+2.82$. The results show that we can successfully predict the loss of a 52B model at 10k steps based on proxy models ranging from 77M to 3.4B, further affirming the accuracy of $\mu$Scaling and usability of \modelname.

\subsection{Efficiency}\label{efficiency and performance}
\paragraph{Cost Ratio.}
We calculate the number of floating point operations (FLOPs) of a model following \cite{DBLP:conf/sc/NarayananSCLPKV21}. Given a Transformer model with $l$ layers, width (\aka hidden size) $w$, sequence length $s$, vocabulary size $V$, and training batch size $B$, the total number of floating-point operations is $M(l,w,s,V,B)=96 B s l w^2\left(1+\frac{s}{6 w}+\frac{V}{16 l w}\right)$. 
Thus, the FLOPs ratio of the whole \modelname process to the training of the full target model can be approximated as:
\begin{equation}
    \frac{M(l,w_1,s,V,B)*t+\sum_{i=2}^{n}M(l,w_i,s,V,B)}{M(l,w_t,s,V,B)},
\end{equation}
where $w_1$ is grid-search width and $t$ denotes corresponding number of grid search trails on $\mu$Transferable HPs (as detailed in Table \ref{tab:muP}). $n$ signifies the count of distinct proxy models used for the fitting loss function, while $w_t$ is the target model width.
In our experiments, the 32-layer GPT comes with a sequence length of 512 and a batch size of 512. The vocabulary contains 100,256 tokens. Thus, the ratio of cost is approximated at 0.131. 
Under similar calculation, the cost ratio for our 64-layer GPT is 0.142.
In short, with just $13.1\%, 14.2\%$ of the one-time pre-training cost, we can predict the loss for 26B and 52B models. Note that without \muscaling, grid-search is needed on the large models, which will enlarge the advantage of our method by several times.

\subsection{Analytical Studies}\label{anal_studies}
\paragraph{Error Comparison with Related Work.} We compare the fitting error of nanoLM with related work that also provided loss numbers, including Cerebras-GPT \cite{DBLP:journals/corr/abs-2304-03208} (in their Table 8), and Pythia \cite{DBLP:conf/icml/BidermanSABOHKP23}. As shown in Table \ref{tab:coeff}, for fitting error, nanoLM achieves slightly better results than related work, even with larger models (26B and 52B). Moreover, the covariances of the fitted coefficients $\{a, b, c\}$ are significantly lower than related work. This indicates that nanoLM is more accurate and confident on its scaling laws, supporting its reliability. 

\paragraph{Impact of Embedding Weights.} We demonstrate an ablation study by Figure \ref{fig:main_result_sanity} and \ref{fig:no_emb}, indicating that our scaling law fits worse if embedding sizes are not counted as model sizes, in contrast to \cite{DBLP:journals/corr/abs-2001-08361,scaling_auto}. This is potentially because $\mu$P concludes that the learning rate of embedding layers should not be scaled down with widths, while most existing methods grid-search for a unified learning rate for all layers on each width, making embeddings learning slower on large models, and the matrix-like parameters dominating the training dynamics. This analysis is conducted with a 12-layer GPT-2 trained on OpenWebText \cite{OpenWeb} for 20k steps.

For other discussions, please see Appendix \ref{appendix: more analy}.

\section{Conclusion and Future Work}
We present \modelname, a cost-efficient benchmark for Large Language Model (LLM) studies, designed to predict losses accurately across various scales without direct training. This benchmark allows for comprehensive comparisons of different model architectures and algorithms, works well with FSDP, and contains curated pre-training datasets ranging from 100B to 2T tokens.
Our loss prediction method, \muscaling, is tested through comprehensive experiments, which demonstrate its effectiveness across different scales, enabling researchers with limited resources to conduct meaningful research for large models. Moving forward, we aim to extend \modelname to more frameworks and tasks (\eg vision), reducing resource waste from non-transferable findings, and enhancing collaboration between academia and industry.

\section*{Impact Statement}
This paper presents work whose goal is to advance the field of Machine Learning. There are many potential societal consequences of our work, none which we feel must be specifically highlighted here.

\bibliography{main}
\bibliographystyle{icml2024}

%%%%%%%%%%%%%%%%%%%%%%%%%%%%%%%%%%%%%%%%%%%%%%%%%%%%%%%%%%%%%%%%%%%%%%%%%%%%%%%
%%%%%%%%%%%%%%%%%%%%%%%%%%%%%%%%%%%%%%%%%%%%%%%%%%%%%%%%%%%%%%%%%%%%%%%%%%%%%%%
% APPENDIX
%%%%%%%%%%%%%%%%%%%%%%%%%%%%%%%%%%%%%%%%%%%%%%%%%%%%%%%%%%%%%%%%%%%%%%%%%%%%%%%
%%%%%%%%%%%%%%%%%%%%%%%%%%%%%%%%%%%%%%%%%%%%%%%%%%%%%%%%%%%%%%%%%%%%%%%%%%%%%%%
\newpage
\appendix
\onecolumn

\section{More analysis} \label{appendix: more analy}
\begin{table}[ht]
    \centering
    \caption{Comparison of Fitted Results: The ``Coeff Values $\&$ Cov" denote the coefficients \{a, b, c\} and the covariance of the power-law function  $y=aC^b+c$.
    }
    \resizebox{\textwidth}{!}{
    \begin{tabular}{@{}c|ccccccccc|ccc@{}}
    \toprule
    \textbf{Model} & \multicolumn{9}{c|}{\textbf{Size in billions $\&$ Loss(Error)}} & \multicolumn{3}{c}{\textbf{Coeff Values $\&$ Cov}} \\ \midrule
    \multirow{2}{*}{Cerebras-GPT} & 0.111 & 0.256 & 0.59 & 1.30 & 2.700 & 6.700 & 13.00 &-&-& 6.76e1 &-8.45e-2 &7.25e1 \\
            & 2.608&2.349&2.181&1.997&1.834&1.704(0.034)&1.572(0.025) &  -&-&4.84e1 & 2.10e-2& 3.44e-1 \\
    \midrule
    \multirow{2}{*}{Cerebras-GPT+$\mu$P} & 0.111 & 0.256 & 0.59 & 1.30 & 2.700 & - & - &-&-& 4.73e1 &-7.37e-2 &5.12e1 \\
            &2.588&2.359&2.155&1.984&1.846(0.004)&-&- &  -&-&1.88e1 & 1.28e-2& 3.05e-1 \\
    \midrule
    \multirow{2}{*}{Pythia} & 0.070&0.160&0.410&1.000&1.400&2.800&6.900&12.00 &-& 9.67e6 &-0.34 &1.42\\
    &2.549&2.204&1.989&1.858&1.889&1.724&1.644(0.049)&1.601(0.019)&- &3.89e7 &8.89e-2& 1.50e-1 \\
    \midrule
    \modelname & 0.077&0.153&0.254&0.381&0.532&0.709&0.911&3.432&5.24e1 & 0.25 &-0.47& 2.82\\
    &3.656&3.389&3.298&3.215&3.198&3.087&3.080&2.958(0.018)&2.883(0.022)& \textbf{7.33e-2} &\textbf{8.50e-2} & \textbf{7.66e-2}  \\
    \bottomrule
    \end{tabular}
    }
    \label{tab:coeff}
\end{table}

\begin{figure*}[h]
  \centering
  \begin{subfigure}{0.45\linewidth}
    \centering
    \includegraphics[width=0.8\linewidth]{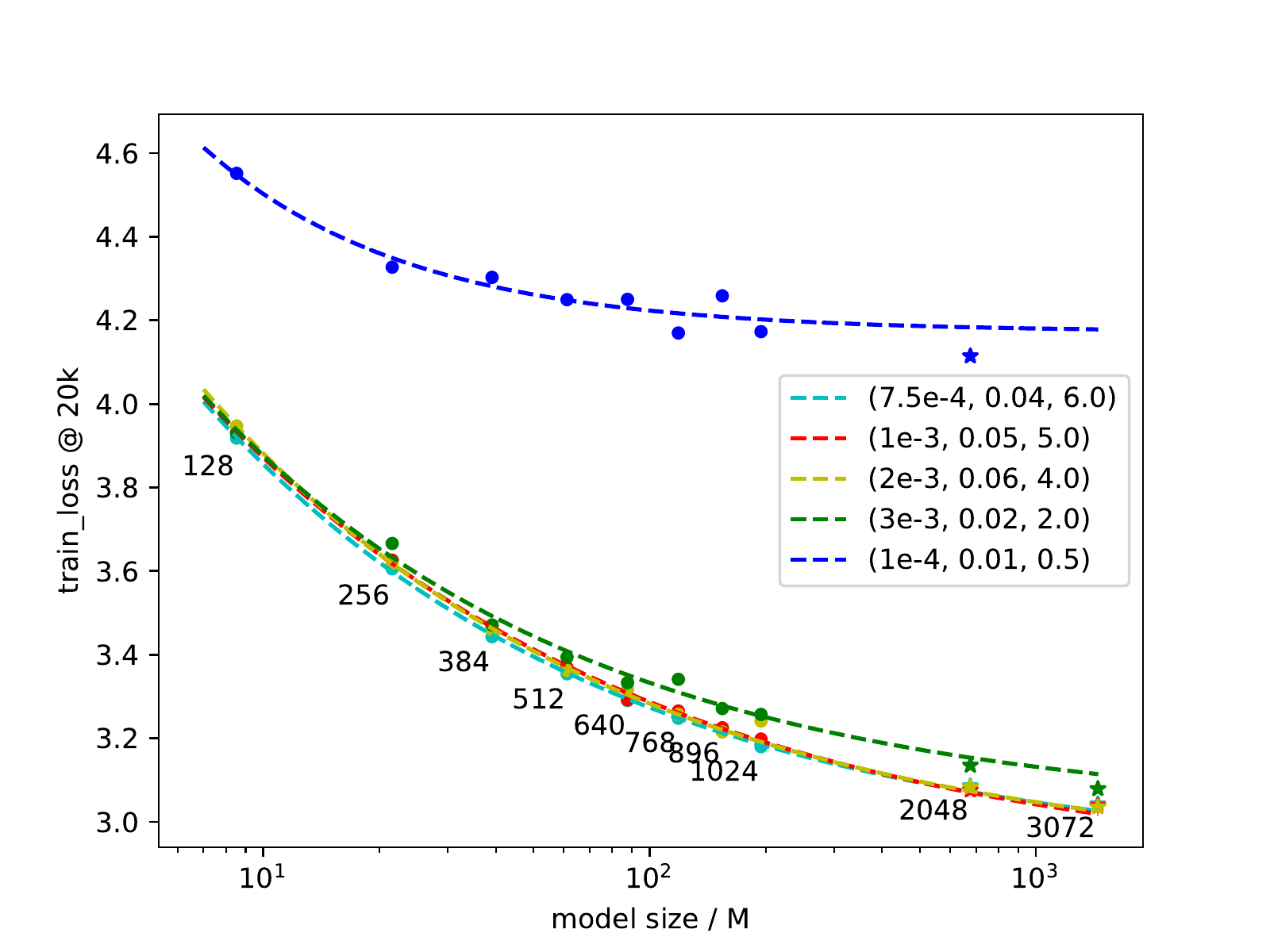}
    \caption{Scaling laws under $\mu$P for different HPs.}
    \label{fig:main_result_sanity}
  \end{subfigure}
  \hspace{0.01\textwidth}
  \begin{subfigure}{0.45\linewidth}
    \centering
    \includegraphics[width=0.8\linewidth]{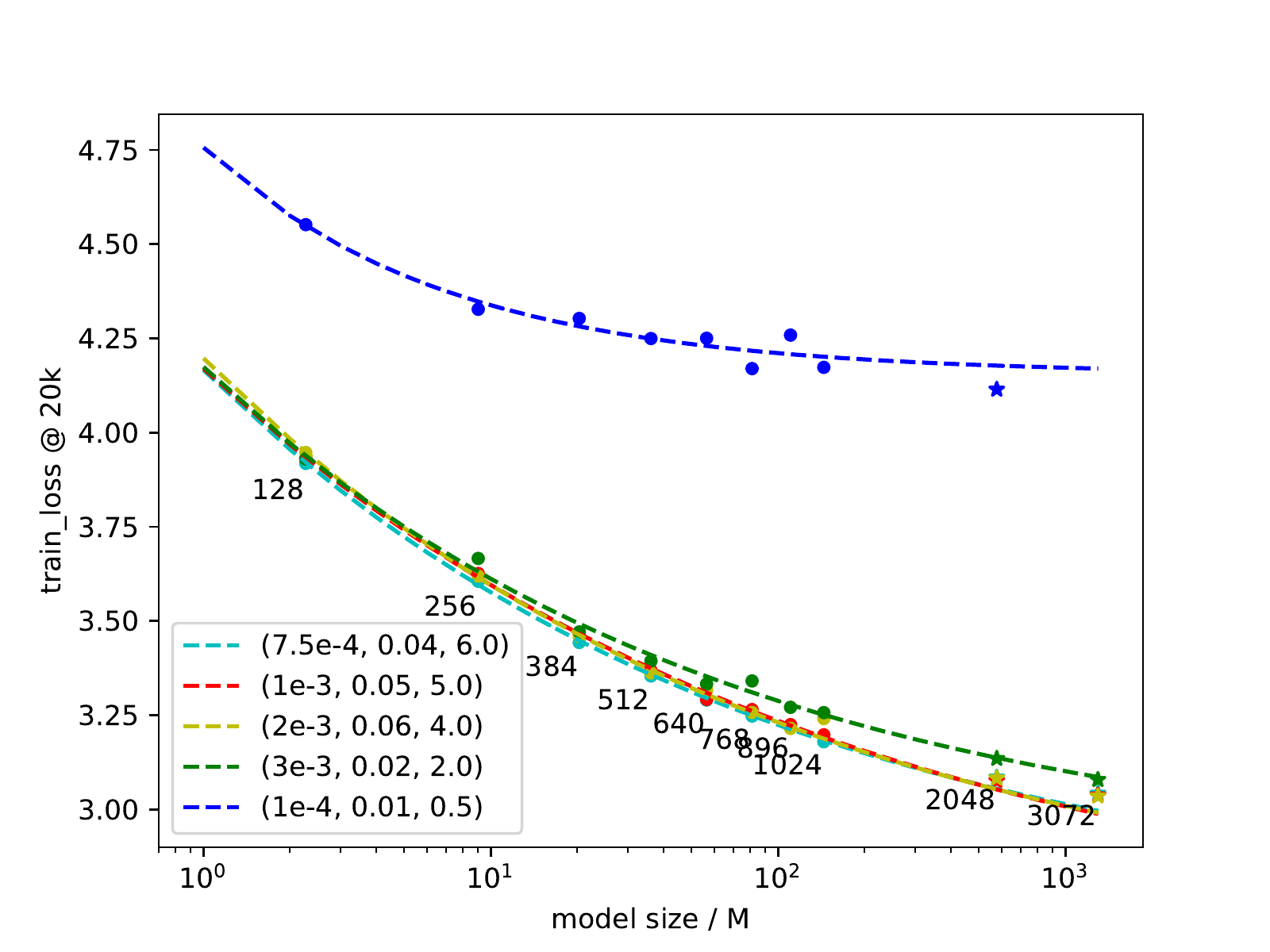}
    \caption{Scaling laws without embedding parameters.}
    \label{fig:no_emb}
  \end{subfigure}
  \hspace{0.01\textwidth}

  \caption{A comparison of \muscaling w/ and w/o embedding size, and the impact of HP loss basins.}
  \label{fig: mup analysis}
\end{figure*}

\paragraph{Scaling law fails outside the loss basins.} Both vanilla scaling laws and \muscaling requires at least one hyperparameter search. We define \textit{loss basin} as an area in the HP space with locally minimal loss where $\partial L / \partial H \approx 0$. Does the scaling law still hold if we randomly pick an $H$ outside the loss basin, and $\mu$Transfer it to all the widths? According to our results in Figure \ref{fig:main_result_sanity}, although ``the wider, the better'' roughly holds as guaranteed by the $\mu$P theory, the numerical scaling law does not trivially generalize well outside the basin (blue and green lines). This observation still exists when models are more sufficiently trained for 20k steps. Thus, we suggest searching for the best HPs first anyway.

\begin{figure}[h]
  \centering
  \begin{subfigure}{0.45\linewidth}
    \centering
    \includegraphics[width=\linewidth]{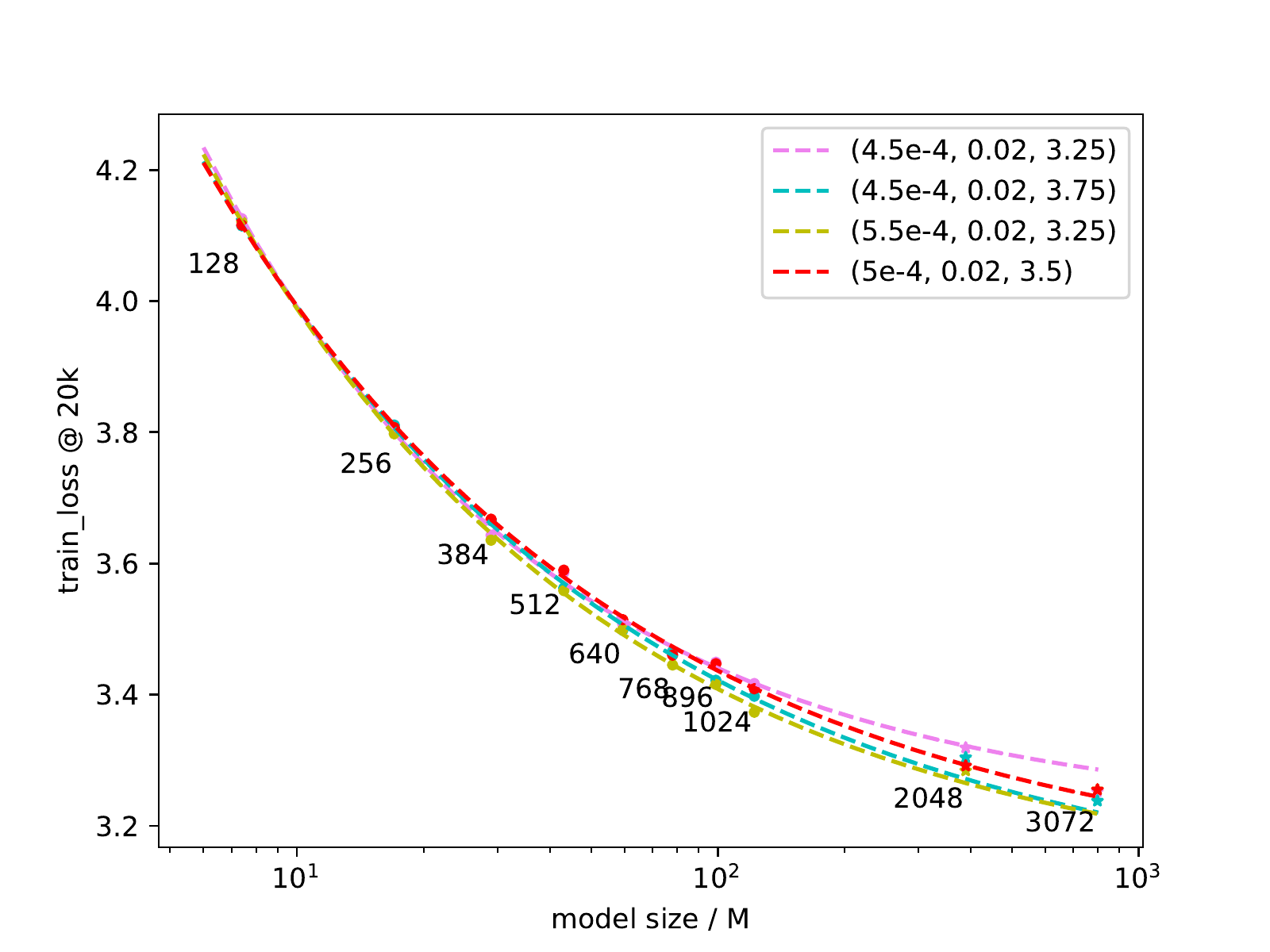}
    \caption{Scaling law for training loss with different
HPs for 6-layer models.}
    \label{fig: smaller_model}
  \end{subfigure}
  \hspace{0.01\textwidth}
  \begin{subfigure}{0.45\linewidth}
    \centering
    \includegraphics[width=\linewidth]{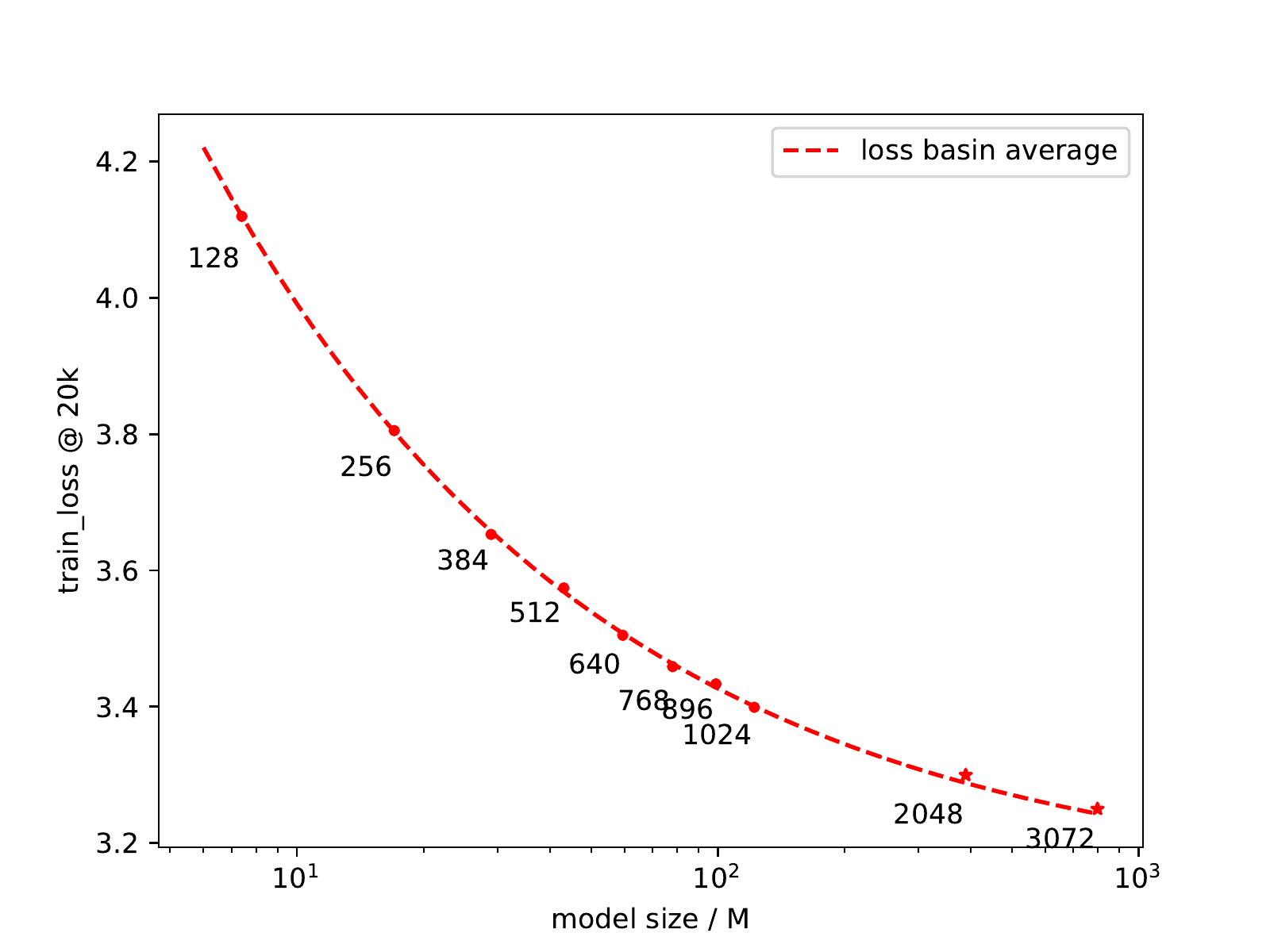}
    \caption{Scaling law for average training loss in the
loss basin of 6-layer models.}
\label{fig: average}
  \end{subfigure}
  \caption{Results with 6-layer Models.}
  \label{fig: 6-layer}
\end{figure}

\paragraph{Take average for extremely small proxy models.}
We find that for extremely small proxy models (\eg GPTs with 256 to 512 width and 6 layers), the fitting results are more vulnerable to slight misalignment of the loss landscapes in $\mu$Transfer. This issue can be addressed by sampling more HP points for each small widths and use the average loss for fitting. Specifically, when we grid-search for the best HPs for 6-layer models with batch size 32 and width 256, we found $(5e-4, 0.02, 3.5)$ being inside the loss basin. As shown in Figure \ref{fig: smaller_model} (red line), \modelname works perfectly for this single point. Then, we explored several other points around it and found that the scaling laws have larger deviations than 12-layer models (Figure \ref{fig: smaller_model}, other lines). We try to balance-off this deviation by fitting scaling laws with the average results across all these HPs near the loss basin. This works perfectly as shown in Figure \ref{fig: average}, and can be practical in applying \modelname because we observe in Figure \ref{fig: smaller_model} that larger widths (\eg 2048, 3072) have significantly lower variances in loss w.r.t different HPs and do not need multiple runs to take average.

\paragraph{General conditions for scaling laws.} Previous scaling laws directly search for HPs on each scale, and the optimal HPs do not satisfy $\mu$P function. It is widely known that these scaling laws still success in a wide range of model sizes. This indicates that $\mu$P is a \textit{sufficient} but not \textit{necessary} condition for scaling laws, and scaling law itself may represent a higher level of universality.

\paragraph{Emergent abilities and inverse scaling law.} We believe that emergent abilities \cite{emergent}  are closely related to evaluation metrics \cite{mirage}, but the pre-training loss and evaluation perplexity based on language model can still be accurately predicted as an effective indicator of model performances. 

\section{Pre-training Data Mix Ratio}
Please see Table \ref{tab:dataset}.

\vspace{-0.5cm}
\begin{table}[h]
    \centering
    \caption{Pre-training data ratio.}
    \begin{tabular}{lrcr}
    \hline Dataset & Sampling prop(\%) & Total tokens(B)  \\
    \hline Arkiv & 6.04 & 28.31  \\
    Books & 5.22 & 24.46  \\
    Falcon RefinedWeb & 20.81 & 97.49 \\
    Falcon RefinedWeb(wiki-like) & 49.78 & 233.21 \\
    OpenWebText2 & 3.11 & 14.59 \\
    StackExchange & 3.81 & 17.84  \\
    Github & 10.18 & 47.70  \\
    Wikipedia & 1.03 & 4.82  \\
    \hline
    \end{tabular}
    \label{tab:dataset}
\end{table}

\section{The Hyperparameter settings for all experiments} \label{appendix: HP settings}
 The specific parameters of the experiment are as follows. 
 (1) The parameters of the model are: vocab\_size = 50304; block\_size = 1024; n\_layer = [12, 32, 64]; head\_size = 64; dropout = 0.0; output\_mult = 1.0; zero\_query = True; zero\_emb = True. hp\_tune\_actual\_width = [128, 256, 384, 512, 640, 768, 896, 1024, 2048, 4096, 8192]; 
 (2) The parameters of the data are: input\_length = 512;  mlm\_probability = 0.15; mean\_noise\_span\_length = 3.0; num\_workers = 2; 
 (3) The parameters of the optimizer are: name = adamwscale;  batch\_size = [16, 512]; total\_steps = [7000, 10000];  warmup\_steps = 5000; lr\_scheduler = cosine; weight\_decay = 0.0; grad\_clip = 1.0;  grad\_acc = 1;  final\_cosine = 1e-5; base\_lr = [5e-4, 1e-3, 5e-3, 1e-2, 3e-2, 5e-2, 7e-2, 1e-1]. 

\section{Grid search results with 256 base width} \label{appendix: GD search}

\begin{table}[h]
    \centering
    \caption{grid search on base width = 256. The specific parameters of the experiment are: n\_layer = 12, batch\_size = 16, hp\_tune\_actual\_width = 256, total\_steps = 7000, initialization std = 0.05, output multiplier = 5.0, base\_lr = [5e-4, 1e-3, 5e-3, 1e-2, 3e-2, 5e-2, 7e-2, 1e-1]. }
    \begin{tabular}{ccccccccccc}
        \toprule
        lr & 5e-4 &1e-3& 5e-3& 1e-2& 3e-2& 5e-2& 7e-2& 1e-1 \\
        \midrule
        12-layer BERT loss  & 7.37 & 7.27 & 5.01 & 4.39 & \textbf{3.9} & 4.17 & 5.24 & 6.97 \\
        12-layer GPT loss  & 7.3 & 7.03 & 5.97 & 5.57 & \textbf{3.74}& 5.86 & 7.22 & 7.25 \\
        12-layer T5 loss  & 6.85 & 6.33 & 5.37 & 5.13 & \textbf{4.71} & 5.14 & 5.28 & 6.45 \\
        \bottomrule
    \end{tabular}
\end{table}

\begin{table}[h]
    \centering
    \caption{grid search on base width = 256. The specific parameters of the experiment are: n\_layer = 64, batch\_size = 512,  hp\_tune\_actual\_width = 256, total\_steps = 10000, initialization std = 0.01, output multiplier = 1.0, base\_lr = [1e-4, 5e-4, 7e-4, 1e-3, 3e-3, 5e-3, 7e-3, 1e-2]. }
    \begin{tabular}{ccccccccccc}
        \toprule
        lr & 1e-4&5e-4 &7e-4&1e-3&3e-3& 5e-3&7e-3& 1e-2 \\
        \midrule
        64-layer GPT loss  & 4.35 & 3.73 & 3.69 & \textbf{3.64} & 8.37 & 13.3 & 9.66 & 8.12 \\
        \bottomrule
    \end{tabular}
\end{table}

\section{Specific Loss Values} \label{appendix: loss value}
\subsection{12-layer Models on C4}

\begin{table}[H]%ht
    \centering
    \caption{training loss on 12-layer@7k steps. The specific parameters of the experiment are: n\_layer = 12, batch\_size = [16, 512],  hp\_tune\_actual\_width = [128, 256, 384, 512, 640, 768, 896, 1024], base\_lr = [1e-3, 3e-2].}
    \label{tab: fitting loss value for 12 layer}
    \begin{tabular}{ccccccccccc}
        \toprule
        width & 128 &256& 384& 512& 640& 768& 896& 1024 \\
        \midrule
        BERT w/o $\mu$P  & 4.25 & 3.71 & 3.59 & 3.52 & 3.47 & 3.42 & 3.37 & 3.40 \\
        BERT with $\mu$P  & 4.45 & 3.74 & 3.63 & 3.56 & 3.49 & 3.47 & 3.44 & 3.43 \\\hline
        GPT w/o $\mu$P  & 4.73 & 4.48 & 4.50 & 4.42 & 4.36 & 4.33 & 4.29 & 4.31 \\
        GPT with $\mu$P  & 4.52 & 4.25 & 4.16 & 4.10 & 4.04 & 4.01 & 4.00 & 3.98 \\\hline
        T5 w/o $\mu$P  & 5.14 & 5.06 & 4.82 & 4.81 & 4.66 & 6.50 & 5.71 & 6.03 \\
        T5 with $\mu$P  & 5.18 & 4.71 & 4.57 & 4.61 & 4.60 & 4.60 & 4.52 & 4.49 \\
        \bottomrule
    \end{tabular}
\end{table}
\subsection{12-layer Models on MC4}

\begin{table}[ht]
\centering
\caption{training loss on 12-layer@20k steps. The specific parameters of the experiment are: n\_layer = 12, batch\_size = [16, 512],  hp\_tune\_actual\_width = [128, 256, 384, 512, 640, 768, 896, 1024], base\_lr = [1e-3, 5e-2].}
\begin{tabular}{ccccccccc}
\toprule
  width & 128 & 256 & 384 & 512 & 640 & 768 & 896 & 1024 \\
\midrule
Bert & 2.79 & 1.36 & 1.24 & 1.17 & 1.14 & 1.10 & 1.08 & 1.06 \\
T5 & 2.15 & 2.05 & 1.76 & 1.62 & 1.54 & 1.51 & 1.45 & 1.41 \\
GPT  & 1.50 & 1.38 & 1.34 & 1.32 & 1.32 & 1.31 & 1.30 & 1.29 \\
Llama  & 1.58 & 1.48 & 1.46 & 1.44 & 1.40 & 1.40 & 1.39 & 1.38 \\
\bottomrule
\end{tabular}
\end{table}

\subsection{32-layer GPT on \modelname Benchmark Data}
\begin{table}[ht]
    \centering
    \caption{training loss on 32-layer@7k steps. The specific parameters of the experiment are: n\_layer = 32, batch\_size = 512,  hp\_tune\_actual\_width = [256, 384, 512, 640, 768, 896, 1024, 2048, 4096, 8192], total\_steps = 7000, base\_lr = 5e-2.}
    \label{tab: fitting loss value for 32 layer}
    \begin{tabular}{ccccccccccc}
        \toprule
        width &256& 384& 512& 640& 768& 896& 1024& 2048 & 8192 \\
        \midrule
        GPT with $\mu$P  & 3.92 & 3.76 & 3.65 & 3.59 & 3.54 & 3.49 & 3.47 & 3.45 &3.41 \\
        \bottomrule
    \end{tabular}
\end{table}

\subsection{64-layer GPT on \modelname Benchmark Data}
\begin{table}[ht]
    \centering
    \caption{training loss on 64-layer@10k steps. The specific parameters of the experiment are: n\_layer = 64, batch\_size = 512,  hp\_tune\_actual\_width = [ 384, 512, 640, 768, 896, 1024, 2048, 8192], total\_steps = 10000, base\_lr = 1e-3.}
    \label{tab: fitting loss value for 64 layer}
    \begin{tabular}{cccccccccc}
        \toprule
        width & 256&384 &512& 640& 768& 896& 1024& 2048 &  8192 \\
        \midrule
        GPT with $\mu$P & 3.656	&3.389	&3.298	&3.215&	3.198	&3.087&	3.080	&2.958	&2.883 \\
        
        \bottomrule
    \end{tabular}
\end{table}
%%%%%%%%%%%%%%%%%%%%%%%%%%%%%%%%%%%%%%%%%%%%%%%%%%%%%%%%%%%%%%%%%%%%%%%%%%%%%%%
%%%%%%%%%%%%%%%%%%%%%%%%%%%%%%%%%%%%%%%%%%%%%%%%%%%%%%%%%%%%%%%%%%%%%%%%%%%%%%%

\end{document}